\pdfoutput=1
\documentclass[11pt]{article}

\usepackage{acl}

\usepackage{times}
\usepackage{latexsym}
\usepackage{flushend}

\usepackage[T1]{fontenc}
\usepackage[utf8]{inputenc}

\usepackage{microtype}

\usepackage{natbib}
\usepackage{graphicx}
\usepackage{amsmath}
\usepackage{bbold}
\usepackage{xcolor}
\usepackage{subcaption}
\usepackage{hyperref}
\usepackage{cleveref}
\usepackage{arydshln}
\usepackage{multirow}
\usepackage{booktabs}
\usepackage[normalem]{ulem}
\useunder{\uline}{\ul}{}
\usepackage{footnote}
\usepackage{bm}
\usepackage{rotating}

\usepackage{caption}

\definecolor{cerulean}{rgb}{0.0, 0.48, 0.65}
\definecolor{asparagus}{rgb}{0.53, 0.66, 0.42}
\definecolor{babyblueeyes}{rgb}{0.63, 0.79, 0.95}
\definecolor{cadmiumorange}{rgb}{0.93, 0.53, 0.18}

\useunder{\uline}{\ul}{}

\title{Grounded Complex Task Segmentation for Conversational Assistants}

\author{Rafal Ferreira, David Semedo, João Magalhães \\
  NOVA LINCS, NOVA School of Science and Technology, Portugal \\
  \texttt{\{rah.ferreira\}@campus.fct.unl.pt, \{df.semedo, jmag\}@fct.unl.pt
 }
}

\begin{document}

\maketitle
\begin{abstract}
Following complex instructions in conversational assistants can be quite daunting due to the shorter attention and memory spans when compared to reading the same instructions. Hence, when conversational assistants walk users through the steps of complex tasks, there is a need to structure the task into manageable pieces of information of the right length and complexity.
In this paper, we tackle the recipes domain and convert reading structured instructions into conversational structured ones. We annotated the structure of instructions according to a conversational scenario, which provided insights into what is expected in this setting.
To computationally model the conversational step's characteristics, we tested various Transformer-based architectures, showing that a token-based approach delivers the best results.
A further user study showed that users tend to favor steps of manageable complexity and length, and that the proposed methodology can improve the original web-based instructional text. Specifically, 86\% of the evaluated tasks were improved from a conversational suitability point of view.\footnote{\url{https://github.com/rafaelhferreira/grounded_task_segmentation_cta}}
\end{abstract}

\section{Introduction}

\begin{figure}[t]
    \centering
    %\captionof{figure}{Caption_2}
    %\label{figu:re}
    \includegraphics[trim={6pt 0 0 0},clip,width=0.95\linewidth]{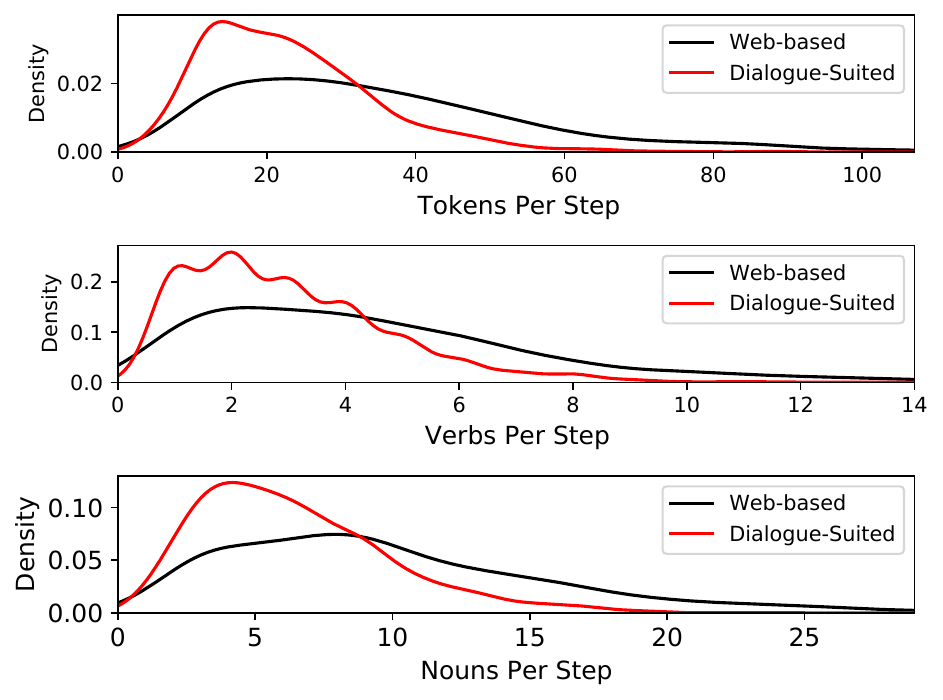}
        \caption{Differences between web-based and dialogue-suited recipes (i.e., ConvRecipes corpus) in terms of the density distribution of tokens, verbs, and nouns per step.}
    \label{fig:web_vs_dialogue_distributions}
\end{figure}

Voice-based assistants can guide users through everyday complex tasks, such as cooking, crafts, and home repairs. These conversational assistants need to understand the users' intention, find a specific recipe, and communicate it in a structured and well-paced manner.
Supporting this type of task-guiding interaction is a recent topic~\cite{taskbot_overview, wizard_of_tasks, task2dial}, where conversational assistants must work hand-in-hand with users in order to guide them throughout the task execution.

We argue that most instructional texts found online are structured in a non-optimal way for conversational assistants, due to the inherent differences between screen and voice-based interfaces. Recipes are a great example where the decomposition of the recipe's text into dialogue-suited steps is critical (example in Appendix~\ref{example_dialog_table}) -- as ~\citet{taskbot_overview} observed, the user is dividing attention through various and possibly parallel actions.
Hence, and following previous knowledge~\cite{short_term_7_2,short_term_4}, we aim for steps that are structured and presented to the user in ordered pieces of information, while dosing complexity, with the aim of achieving an efficient task completion. 

To tackle this new problem, we part ways with topic-based segmentation methods~\cite{wiki_727,choi_dataset,wikisection_sector} and propose a novel human-focused methodology to convert reading-structured instructions into conversational ones.
Figure~\ref{fig:web_vs_dialogue_distributions} offers a clear view of the differences between the original web-based recipes and their dialogue-suited counterparts. The distribution of linguistic characteristics such as length, verbs (which cover actions), and nouns (covering ingredients, tools, etc.), confirms that dialog-suited instructions should avoid overwhelming users' short-term memory~\cite{short_term_7_2,short_term_4}.

Our proposed methodology starts with the curation of a corpus which we call \textit{ConvRecipes}, where online recipes are segmented into recipes with steps more suited to a conversational agent.
Moreover, we identify the key traits of a conversational step. 
An example of this can be seen in Figure~\ref{fig_web_dialog_recipe}.
This example shows the need for models that can tackle our task, in specific, in ''Step 2'' of the task, where it is noticeable that the step would be very difficult to follow in a conversational assistant due to the long sentence and the inherent complexity of its actions.

To tackle this task computationally, we propose the Dialogue-Task Segmenter Transformer (\textit{DTS-Transformer}), which follows state-of-the-art approaches in text-segmentation~\cite{text_segmentation_bert,transformer_squared,meetings_segmentation} and adopts a Transformer-based backbone~\cite{vaswani_attention}. Distinct from previous work, we follow a token-level approach which by modeling steps' text at a finer granularity, is capable of better modeling the inherent structural characteristics of conversational tasks. 
Note that, we did not follow generative approaches and ground our task segmentation task on the recipes' original text. We do this to avoid the risk of introducing hallucinations or mistakes in step-by-step procedures~\cite{wizard_of_tasks}.

Finally, we validated the proposed methodology with automatic experiments, and, more interestingly, with a user evaluation. We observed that the best \textit{DTS} model, a \textit{T5-3B} Encoder backbone, trained on the proposed ConvRecipes corpus, was able to improve the conversational structure of 86\% of the evaluated tasks. This evidences both the conversational characteristics of the ConvRecipes corpus and the effectiveness of the model's approach to the grounded conversational task segmentation task.

Next, we will relate our contributions to previous corpus and methods. In Section~\ref{sec_dataset}, we carefully detail the proposed methodology. Experimental validation and user evaluations are presented in Section~\ref{sec_experiments}, and we conclude with the final takeaways and future work.

\section{Related Work}
\paragraph{Related Corpora.}
While conversational-suited task segmentation is a novel task, multiple datasets have been created to address article-based text segmentation, with the earliest ones being the Choi Dataset~\cite{choi_dataset}, where each document is represented by the concatenation of 10 random passages from a large corpus, and the 
RST-DT dataset~\cite{RST_DT_dataset}, which focuses on intra-sentente granularity on Wall Street Journal articles.
Topic and document-section-oriented segmentation datasets such as 
Wiki-727~\cite{wiki_727} and WikiSection~\cite{wikisection_sector} are comprised of Wikipedia articles and focus on topic and section-based text segmentation.
Closer to our domain, we highlight works with instructional text such as Task2Dial~\cite{task2dial} and the Wizard of Tasks~\cite{wizard_of_tasks}, which rewrite the tasks' text into dialogue-suited steps.
Our approach focuses on grounded structuring of task instructions for dialog while avoiding hallucination problems common in generative/re-writing approaches.
We also take a step further by identifying the fundamental traits of conversational-suited tasks, in a principled manner. 

\begin{figure}[t]
	\begin{minipage}{0.97\linewidth}
		%\caption{Caption }
		%\label{tab:le}
		\small
		\centering
		\resizebox{\textwidth}{!}{%
		\begin{tabular}{p{\linewidth}}

\large \textbf{Title:} Baked Bananas Recipe   \\ \toprule
\hspace{2.5cm}\textbf{Web-based Recipe}       \\ \toprule
\begin{tabular}[t]{@{}p{\linewidth}@{}}
\textbf{Step 1:} \textcolor{cerulean}{Preheat} \textcolor{cadmiumorange}{oven} to 190 \textcolor{cadmiumorange}{degrees} C. \textcolor{cerulean}{Spray} a \textcolor{cadmiumorange}{baking dish} with \textcolor{cadmiumorange}{cooking spray}. 
\\
\textbf{Step 2:}
\textcolor{cerulean}{Arrange} \textcolor{cadmiumorange}{banana halves} in the \textcolor{cerulean}{prepared} \textcolor{cadmiumorange}{baking dish}. \textcolor{cerulean}{Drizzle} \textcolor{cadmiumorange}{maple syrup} over \textcolor{cadmiumorange}{bananas} and top with \textcolor{cadmiumorange}{cinnamon}. \textcolor{cerulean}{Bake} in the \textcolor{cadmiumorange}{oven} until heated through, 10-15 \textcolor{cadmiumorange}{minutes}.\end{tabular} \\ \toprule
\hspace{2.5cm}\textbf{Dialogue-suited Recipe} \\ \toprule
\begin{tabular}[t]{@{}p{\linewidth}@{}}
\textbf{Step 1:} \textcolor{cerulean}{Preheat} \textcolor{cadmiumorange}{oven} to 190 \textcolor{cadmiumorange}{degrees} C. \\ \textbf{Step 2:} \textcolor{cerulean}{Spray} a \textcolor{cadmiumorange}{baking dish} with \textcolor{cadmiumorange}{cooking spray}.\\ \textbf{Step 3:} \textcolor{cerulean}{Arrange} \textcolor{cadmiumorange}{banana halves} in the prepared \textcolor{cadmiumorange}{baking dish}. \textcolor{cerulean}{Drizzle} \textcolor{cadmiumorange}{maple syrup} over \textcolor{cadmiumorange}{bananas} and top with \textcolor{cadmiumorange}{cinnamon}.\\ \textbf{Step 4:} \textcolor{cerulean}{Bake} in the \textcolor{cadmiumorange}{oven} until heated through 10-15 \textcolor{cadmiumorange}{minutes}.
\end{tabular} \\ \bottomrule
\end{tabular}}
	\end{minipage}\hfill
	\caption{Example of conversion from web/reading-based format to a dialogue-suited format. In blue and orange, we highlight the verbs and nouns, respectively.}
	\label{fig_web_dialog_recipe}
\end{figure}

\paragraph{Methods and Models.}
Initial works for text segmentation were mostly based on statistical and unsupervised approaches, such as TextTiling~\cite{text_tiling} and C99~\cite{choi_dataset}. After these, supervised neural methods, particularly with the use of RNNs were utilized. In~\cite{attention_rnn_cnn}, a CNN is used to generate sentence embeddings in conjunction with an LSTM to keep sequential information. \citet{segbot} also presents an RNN-based model with an additional pointing mechanism and in~\cite{wiki_727} it is used a hierarchical Bi-LSTM model.

Currently, the state-of-the-art is based on supervised Transformer-based 
approaches~\cite{text_segmentation_bert,transformer_squared,meetings_segmentation}. In~\cite{text_segmentation_bert}, cross-segment and hierarchical models are proposed, where predictions are made based on consecutive segments or sentence-based representations of the segments. \citet{transformer_squared} presented a hierarchical approach combining sentence and cross-segment embeddings.~\citet{bilstm_with_transformer} proposed a hierarchical BiLSTM to complement BERT's~\cite{bert_original} sentence representations, aided by a coherence-related auxiliary task. Some approaches such as~\cite{organizing_tasks_kiseleva}, tackle task structuring as a generation task, where an end-to-end pipeline is proposed to generate day-to-day tasks. In a dialogue setting, \citet{meetings_segmentation} applied a BERT model for transcript-based meetings segmentation~\cite{icsi_meetings_dataset, ami_meeting_corpus} by calculating the similarity between segment embeddings given by a pre-trained model.
Given the particular intricacies of conversational-suited task structuring, while we also adopt a Transformer backbone, we propose a task segmentation model that makes decisions at a token-level being able to consider the global task's structure.

\section{Structuring Conversational Tasks}
\label{sec_dataset}
Our hypothesis is that the recipe instructions found online are not suited for conversational assistants, motivating both the task and the dataset collection efforts. 
To convert instructions from a reading structure into a conversationally structured format, we followed a human-focused methodology. First, we collected task instructions and ran a user study to curate them as conversational instructions. Second, we ask users to annotate the relevance of various conversational instructions traits. Third, we analyzed the linguistic characteristics of reading instructions compared to conversational task instructions. Finally, we modeled conversational-steps computationally with various Transformer-based~\cite{vaswani_attention} architectures.

\subsection{A Conversational-Tasks Corpus}
\label{sub_annotation_dataset}
Currently, there are no explicit corpora for studying the grounded segmentation of a recipe into conversational-suited steps.
The closest examples are either section-based document segmentation~\cite{wiki_727,wikisection_sector} or rewriting/generative approaches~\cite{wizard_of_tasks, task2dial} which are prone to hallucinations.
In this section, we introduce the methodology used to create the  \textit{ConvRecipes} corpus, consisting of recipes segmented into conversational-suited steps.

\subsubsection{Tasks Collection and Annotation}
\label{sub_sec:data_collection}
To create the ConvRecipes corpus, we collected recipes from a popular recipes website, where each recipe is self-contained and composed of various steps in English with arbitrary lengths.
We started by filtering out recipes with fewer than three steps due to having a structure that is too simple.
After this, near-duplicate recipes were identified with SimHash~\cite{simhash} and removed.

\vspace{2mm}
\noindent
\textbf{Conversation-Steps Annotation.}
Even though recipes are human-edited, we argue that they are written for a reading-based setting, making them ill-suited to be used in a conversational setting.
Hence, to create grounded conversational instructions, we conducted a user study. 
In total, we had 8 annotators, 6 male and 2 female all Computer Science MSc. and or Ph.D. students. All annotators had experience with both conversational assistants and cooking applications, making them particularly suited for this annotation task.

The annotators were shown the original recipes and asked to propose (or not) changes to make the recipes dialog-suited, either by adding and/or removing steps.
Figure~\ref{fig_web_dialog_recipe} illustrates the annotation process process: given a recipe formatted for the Web, the goal is for the annotator to identify the structure that is better suited for a conversational setting. 
This approach makes the segmentation grounded on the original task, avoiding the introduction of mistakes prone to happen when using rewriting approaches.

\subsubsection{On the Traits of a Conversational Step}
\label{sub_sub_traits}
After the annotation process described in the previous section, the annotators were asked to quantify, on a Likert scale of 1 to 5, the importance of various conversational traits. In particular, we considered: \textit{Complexity}, \textit{Clarity}, \textit{Length and \#Steps}, \textit{Ability to Parallelize Tasks}, and \textit{Naturalness}.
For the exact description of these traits refer to Appendix~\ref{app_traits_description}.
This evaluation of the traits aims to further inform us what users value in this conversational task-guiding assistance setting~\cite{taskbot_overview}.

Table~\ref{tab_trait_importance} shows the results of the analysis of the traits. The results reveal that although all traits have some importance, users mostly focus on the complexity and length of the steps, which are generally connected with each other. This means that managing complexity and ensuring a balance in the information given to the user is paramount.
On the other hand, the naturalness and the ability to perform parallel tasks were considered less important traits, which seem to indicate that users are not so concerned with language naturalness given that the step should be short and not too complex.

\begin{table}[tbp]
\centering
%\resizebox{\linewidth}{!}{%
\begin{tabular}{lc}
\toprule
\textbf{Conversational-Step Trait}      & \textbf{Importance} \\ \midrule
(1) Complexity                   & 4.5                 \\
(2) Step Length \& \#Steps       & 4.2                \\
(3) Clarity                      & 3.8                \\
(4) Naturalness                  & 3.6                \\
(5) Ability to Parallelize Tasks & 3.4                \\ \bottomrule
\end{tabular}%
%}
\caption{Trait importance on a 1 to 5 scale. A higher value represents higher importance.}
\label{tab_trait_importance}
\end{table}

\begin{table}[t]
\centering
%\resizebox{\linewidth}{!}{%
\begin{tabular}{lrr}
\toprule
                    & \textbf{Reading} & \textbf{Dialog} \\ \midrule
Avg. \# Tokens      & \multicolumn{2}{c}{135}                 \\
Avg. \# Sentences   & \multicolumn{2}{c}{9.3}                \\ \hdashline
Avg. \# Steps       & 3.80               & 5.85               \\
Avg. \# Tokens step & 35.44              & 23.03              \\
Avg. \# Sents. step & 2.44               & 1.59               \\
Avg. \# Verbs step  & 4.23               & 2.75               \\
Avg. \# Nouns step  & 9.92               & 6.44               \\ \bottomrule
\end{tabular}%
%}
\caption{Comparison between the 300 original reading-based recipes and the manually annotated set.}
\label{tab_original_vs_annotated}
\end{table}

\subsection{ConvRecipes Corpus Analysis}
After preparing and curating the task instructions, we analyzed and compared the original to the curated data in order to understand how the language differs from a web/reading setting to a conversational setting.

\subsubsection{Reading-suited vs Dialog-suited}
In total, 300 recipes were annotated, where 59 recipes were left without changes, and the remaining 241 (80.3\%), had at least one new step added, with one, two, and three or more breaks added 75, 66, and 47 times, respectively. 
Only one recipe was annotated with fewer steps than the original.
This result shows that the reading-based instructions are not optimal for a conversational setting, generally missing critical segmentations. Table~\ref{tab_original_vs_annotated} further evidences the difference between the original and the conversational-suited instructions, where it is clear that there is a preference for shorter segments with fewer actions.
These results correlate with the importance of the conversational traits (Table~\ref{tab_trait_importance}), which showed that the complexity and number of steps are particularly important in this setting.
Thus, ConvRecipes presents a step forward in discovering the optimal structure for instructional text in a conversational scenario.

\subsubsection{Linguistic Style of Conversational-Steps}
\label{sub_sec:dataset_statistics}
\begin{figure}[t]%
    \centering
    {{\includegraphics[width=0.49\linewidth]{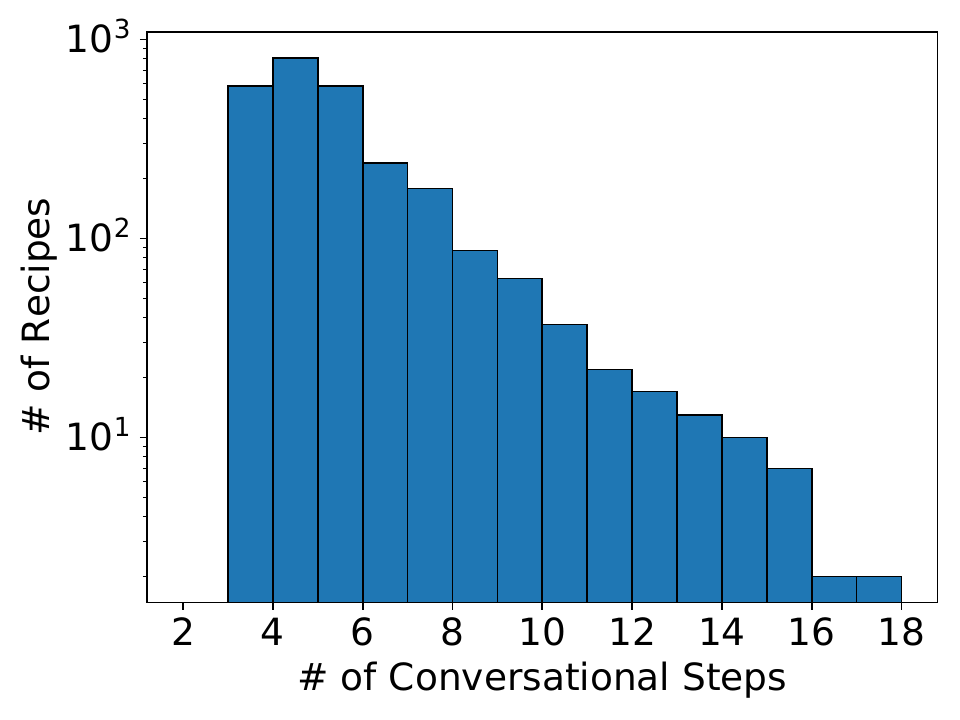} }}%
    %\qquad
    {{\includegraphics[width=0.49\linewidth]{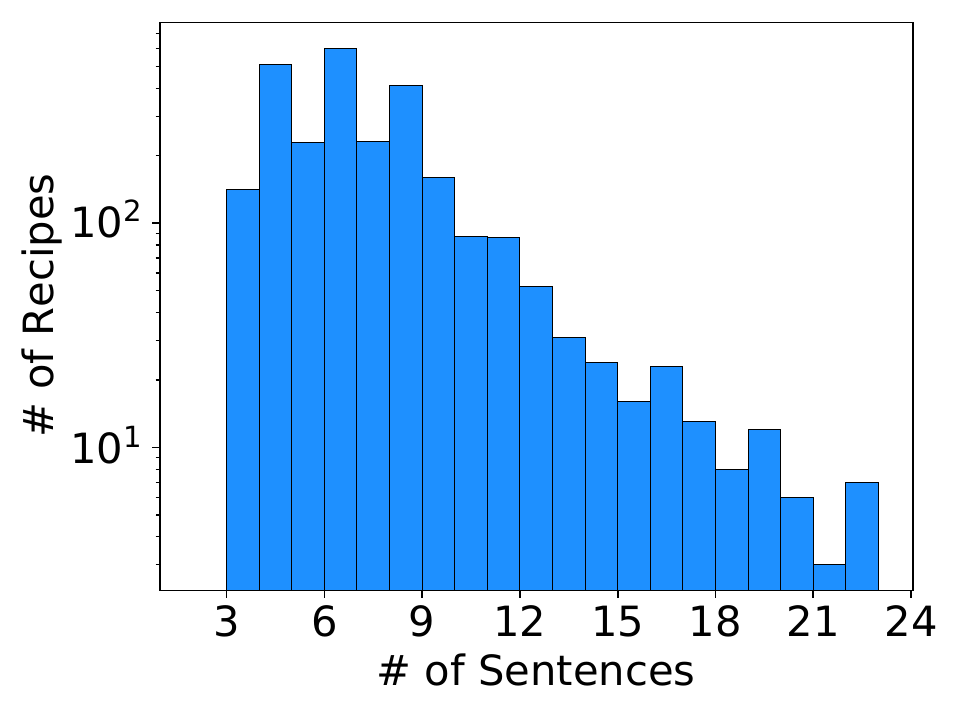} }}%\\
    {{\includegraphics[width=0.49\linewidth]{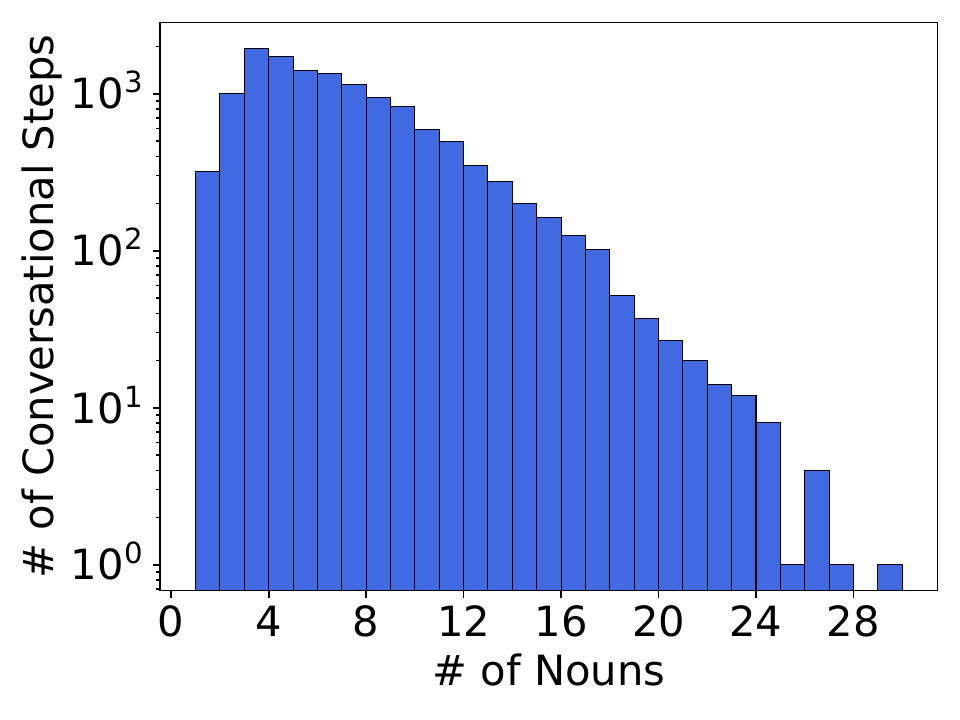} }}%
    %\qquad
    {{\includegraphics[width=0.49\linewidth]{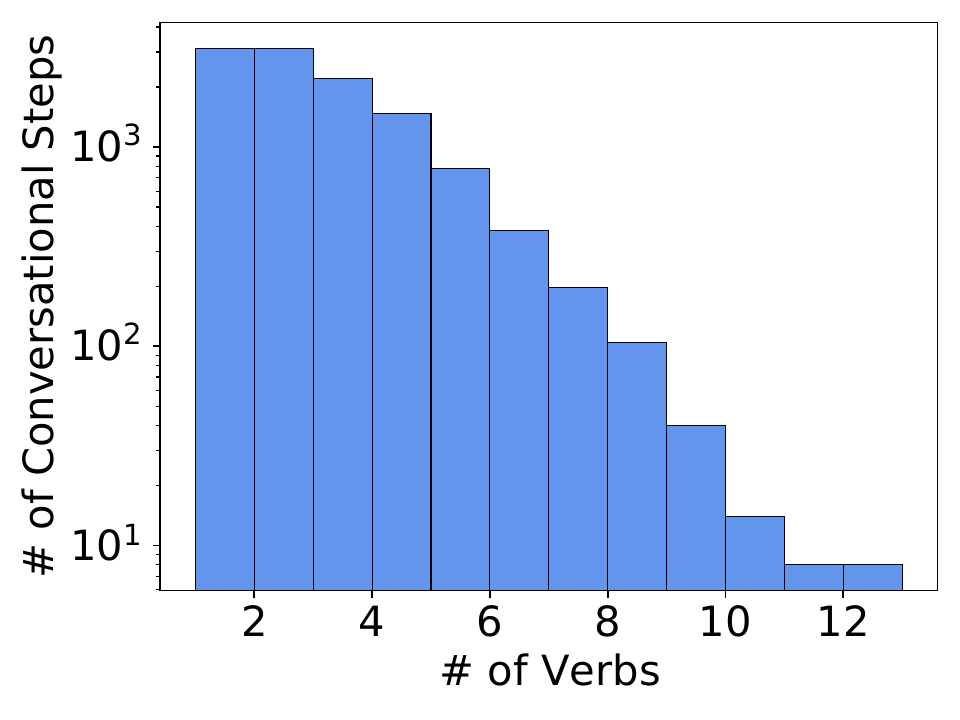} }}%
    \caption{ConvRecipes statistics: conversational-steps per task (top-left), sentences per task (top-right), nouns per conversational-step (bottom-left), and verbs (bottom-right) per conversational-step.}%
    \label{fig_full_recipe_stats}%
\end{figure}

Figure~\ref{fig_full_recipe_stats} shows the corpus's distribution of conversational steps, sentences, nouns, and verbs. The figure indicates that there is a lot of variability that needs to be correctly addressed, due to each recipe having a particular structure.

\begin{figure*}[t]%
    \centering
    {{\includegraphics[width=0.47\linewidth]{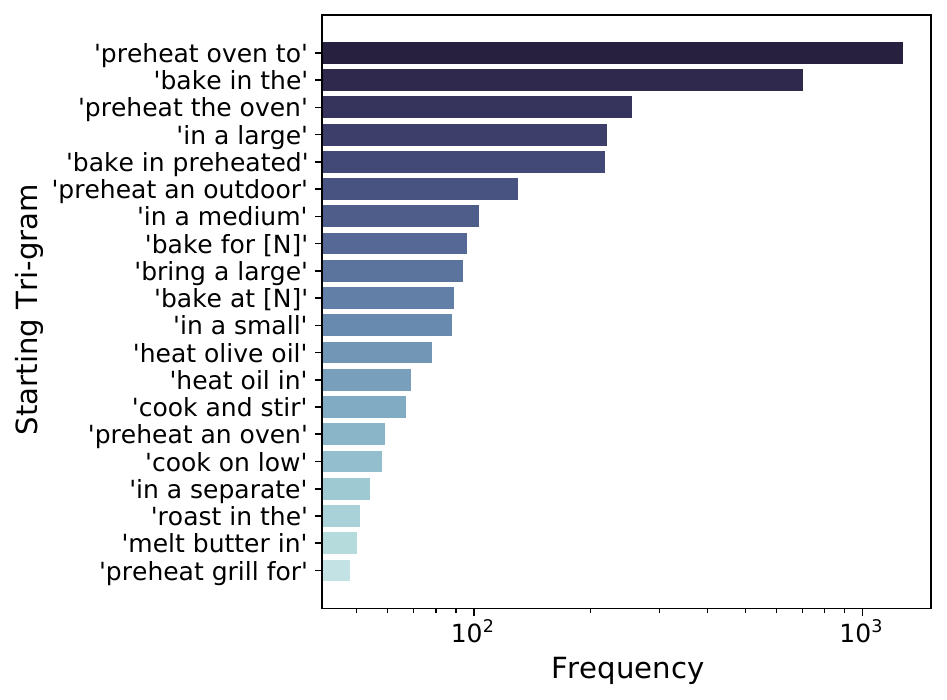} }}%
    \qquad
    {{\includegraphics[width=0.47\linewidth]{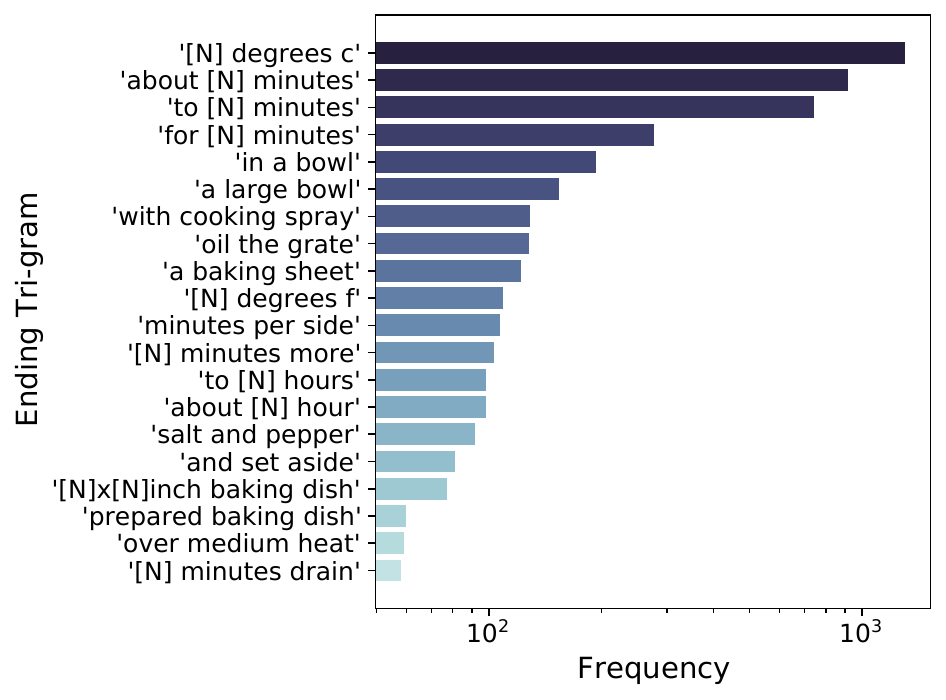} }}%
    \caption{Distribution of the top-20 most frequent starting (left) and ending (right) tri-grams.}%
    \label{fig_top_20_trigrams}%
\end{figure*}

In contrast to other corpora~\cite{wiki_727, choi_dataset}, ConvRecipes is written in an instructional/imperative format, using actionable verbs mostly related to the cooking domain such as ``stir'', ``bake'' and ``mix''.
Analyzing how steps start and end, can also bring some insights into the segmentation behavior, so we examined the most common starting and ending n-grams of each step.
The top-20 starting and ending tri-grams are available in Figure~\ref{fig_top_20_trigrams}. 
This showed that many of the steps have temperature mentions, e.g. ``preheat oven to'', or time aware mentions \textit{e.g.}, ``for [N] minutes'' ([N] is a placeholder replacing the number). 
These indicate a start/end of an action which in turn reflects a new step.
It is important to note that the majority of both bi-grams (65\%) and tri-grams (80\%) are only used once, which shows the diversity of actions available, creating a more complex challenge for data-driven approaches.

\subsubsection{Corpus Processing.}
Annotating a large number of recipes is labor-intensive and expensive. 
Thus, we use the 300 manually annotated recipes as the test set and create the training and validation splits automatically, by using the average number of sentences per step of the annotated set (1.59, Table~\ref{tab_original_vs_annotated}) as a maximum threshold to choose recipes from a non-annotated set. 
These non-annotated recipes use the original step information as the ground-truth step labels, in a similar way as the section markers in the Wiki-727 dataset~\cite{wiki_727}. 
This resulted in a dataset with 1930, and 424 recipes for training and validation, respectively. As mentioned before, the test set is composed of the 300 annotated recipes.

To conclude, the aim of this corpus is to create models that learn how to identify steps and segment a task into grounded dialogue-suited steps. Hence, we concatenate all the steps together, resulting in an unstructured text with no segment-identifying structure.
Step annotations are then used as labels to train and evaluate the models. 

\subsection{Dialog-Task Structuring Transformer}
\label{sub_model_formalization}
To learn the structure of a conversational task, we processed the entire task's text as a whole.
By explicitly receiving the entire input sequence, we aim to take into account the size and position of each segment token relative to all the other tokens.
Consequently, with a single pass over the input, this approach is able to output all segment predictions, making it more efficient than sentence-based embedding models that output a prediction per sentence~\cite{text_segmentation_bert}. 

Given the characteristics of the Transformer model~\cite{vaswani_attention}, we use it as the basis for our \textit{Dialogue-Task Structuring} (DTS) model. In particular, we feed the model with the complete recipe, allowing it to create contextualized token representations of the entire recipe. This allows the model to consider all of the tokens in the recipe via the self-attention mechanism. 

After the input has been processed by the Transformer, we apply a binary segment-break prediction head, i.e., a linear layer followed by a \textit{softmax} to the embedding of each segment identifying token ($emb_t$), outputting the probability of a token ($t$) being a \textit{segmentation token}:
\begin{equation}
    P_{seg}(t_i) = softmax(FFNN(emb_{[t_i]})),
    \label{eq_model_input}
\end{equation}
To identify these end-of-segment tokens, generally punctuation marks (\textit{e.g.} ``.'', ``!'', ``;''), we use Spacy~\cite{spacy} to perform basic sentence segmentation over the recipe's text.
Finally, we train the model using the cross-entropy loss between the model predictions, $\hat{y}$, and the binary segmentation labels, $y$, as the following:
\begin{equation}
    L_{CE} = y \cdot \log{\hat{y}} + (1-y) \cdot \log{(1-\hat{y})},
    \label{eq_cross_entropy}
\end{equation}

\section{Experiments}
\label{sec_experiments}
In this section, we demonstrate how the proposed framework tackles the challenge of structuring task instructions in a conversational setting. Experimental validation was done with both automatic metrics and human evaluation. 

\subsection{Metrics}
 
We use Precision, Recall, and F-score to measure the detection of the correct location of a conversation step following~\citet{segbot}.
Moreover, we followed~\cite{wiki_727,wikisection_sector}, and used the text segmentation metric $P_k$~\cite{pk_metric}, which compares the predicted segmentation with the ground-truth labels, where a lower value represents a better model.

\subsection{DTS Backbones and Implementation Details}
\label{sub_sec:implementation}
As the basis for our models, we used pre-trained Transformer models.
We tested with the encoder-only model BERT~\cite{bert_original}, the encoder-decoder model T5~\cite{t5_model} in both an encoder-decoder (Enc-Dec) setting and in an encoder-only (Enc-only), i.e. decoder is not used.
When using an E-D model, the input sequence of the decoder is the same as the encoder as in~\cite{bart_model} for the extractive QA task (i.e. there is no actual decoding).
To identify the candidate segments, we used Spacy~\cite{spacy}, to be more robust than a simple punctuation-based approach.
We evaluated in the test set the model with the best performance in the validation set in terms of F-Score.
Additional information about model training is provided in Appendix~\ref{app_implementation_details}.

\subsection{Baselines}
As baselines, we considered random and uniform approaches, a classic method~\cite{text_tiling}, and a strong baseline based on a cross-encoder~\cite{text_segmentation_bert}:

\vspace{1.5mm}
\noindent
\textbf{Rand$_p$} and \textbf{Every$_n$}: unsupervised methods which use Spacy~\cite{spacy} to identify sentences. $p$ is the probability of breaking at each sentence, and $n$ is the number of consecutive sentences to break.

\vspace{1.5mm}
\noindent
\textbf{TextTiling}~\cite{text_tiling}: one of the earliest text segmentation methods based on lexical co-occurrence.

\vspace{1.5mm}
\noindent
\textbf{Cross-Segment (CrossSeg)}~\cite{text_segmentation_bert}: BERT-Base~\cite{bert_original} model with a classification head 
that predicts if a pair of input sentences should be segmented.

\subsection{Results and Discussion}
In Table~\ref{tab_results_summary_2}, we present the results of the baselines, along with the results of the proposed DTS models.

\newcommand{\spheading}[2][8em]{% \spheading[<width>]{<stuff>}
  \rotatebox{90}{\parbox{#1}{\centering #2}}}
\begin{table*}[tbp]
\centering
%\resizebox{\linewidth}{!}{%
\small
\begin{tabular}{clccccc}
\toprule
&\textbf{Model}        & \textbf{\# Params} & \textbf{P$_k$$\downarrow$}       & \textbf{Precision$\uparrow$}              & \textbf{Recall$\uparrow$}            & \textbf{F1$\uparrow$}           \\ \midrule
\multirow{6}{*}{\spheading{\small\textbf{\ \ \ \ \ Baselines}} }
& Rand$_{0.5}$         
& -                  & 35.4 $\pm$ 0.3        & 59.9 $\pm$ 0.5          & 49.7 $\pm$ 0.8        & 51.7 $\pm$ 0.6        \\
& Rand$_{0.75}$         & -                  & 28.3 $\pm$ 0.5        & 61.2 $\pm$ 0.4          & 75.0 $\pm$ 0.9          & 65.2 $\pm$ 0.6        \\
& Every$_1$             & -                  & 23.3          & 60.9            & \textbf{98.8} & 73.8          \\
& Every$_2$             & -                  & 37.9          & 59.6            & 37.9          & 44.9          \\ 
& TextTiling            & -                  & 28.4          & 58.7            & 67.7          & 61.4          \\
& CrossSeg         & 110 M              & 19.5 $ \pm$ 0.4       & 77.5 $ \pm$ 0.9         & 79.5 $ \pm$ 1.6       & 76.5 $ \pm$ 0.4       \\
\multirow{8}{*}{\spheading{\small \textbf{Dialogue Task Segmenter (DTS)}} }
%\multirow{5}{*}{\begin{turn}{90}{\footnotesize \textbf{Dialog Task Segmenter}}\end{turn}} 
& BERT-Base (All*)      & 110 M              & 22.5 $\pm$ 0.3        & \textbf{93.4 $\pm$ 0.1} & 58.7 $\pm$ 0.4        & 69.6 $\pm$ 0.4        \\ \cmidrule{2-7}
& BERT-Base             & 110 M              & 19.1 $\pm$ 0.4        & 75.8 $\pm$ 0.7          & 83.6 $\pm$ 0.7        & 77.5 $\pm$ 0.4        \\ 
& BERT-Large            & 340 M              & 18.4 $\pm$ 0.2        & 77.0 $\pm$ 1.7            & 83.6 $\pm$ 2.8        & 78.1 $\pm$ 0.5        \\
& T5-Base (Enc-only)             & 110 M              & {\ul 17.7 $\pm$ 0.2}  & 77.9 $\pm$ 0.7          & 84.2 $\pm$ 0.5        & {\ul 79.0 $\pm$ 0.1}    \\
& T5-Base (Enc-Dec)           & 220 M              & 18.1 $\pm$ 0.6        & 77.9 $\pm$ 0.3          & 82.9 $\pm$ 1.6        & 78.5 $\pm$ 0.8        \\
& T5-Large (Enc-only)            & 335 M              & 18.1 $\pm$ 0.2        & 77.4 $\pm$ 0.4          & 84.1 $\pm$ 0.4        & 78.6 $\pm$ 0.3        \\
& T5-Large (Enc-Dec)          & 770 M              & {\ul 17.7 $\pm$ 0.2}  & {\ul 79.1 $\pm$ 0.8}    & 81.9 $\pm$ 0.9        & 78.5 $\pm$ 0.2        \\
& T5-3B (Enc-only)               & 1.5 B              & \textbf{17.0 $\pm$ 0.4} & 78.3 $\pm$ 1.0            & {\ul 85.9 $\pm$ 0.9}  & \textbf{80.0 $\pm$ 0.2} \\ \bottomrule
\end{tabular}%
%}
\caption{Results on the ConvRecipes's test set from an average of 3 runs per model. 
\textit{All*} \\ indicates that the model was trained on the set of all recipes crawled, in their original form.}
\label{tab_results_summary_2}
\end{table*}

\paragraph{Importance of Conversational-Aware Corpora.}
We trained the same DTS model with a BERT-Base backbone: one on all crawled raw recipes (20.000 recipes), identified as \textit{(All*)}, and one on the ConvRecipes training set (\textit{BERT-Base}).
The results on Table~\ref{tab_results_summary_2}, show that \textit{BERT-Based (All*)} obtained the highest Precision (93.4), since its training samples have fewer breaks, the model makes less, but correct, break predictions. On the other hand, it achieved the lowest Recall of all supervised methods.
More importantly, the results clearly show the importance of training models with suited data, yielding a $\bm{P}_k$ relative improvement of $15\%$ (\textit{Bert-Base}). This result indicates that the ConvRecipes dataset is constructed in a way that embeds the traits of conversational task instructions (Section~\ref{sub_sub_traits}).  

% unspupervised baselines
\paragraph{General results.}
In Table~\ref{tab_results_summary_2}, we observe that 
%Regarding the unsupervised baselines, 
the baselines \textit{Rand$_p$}, \textit{Every$_n$} and \textit{TextTiling} do not generally break the steps at the correct locations as indicated by their low precision ($\leq$ 62\%). However, since it implicitly enforces a step distribution that resembles the dataset, \textit{Every$_1$} achieves a fairly good $\bm{P}_k$, while also achieving a recall close to $100\%$ due to breaking at every sentence (it is $\neq 100\%$ due to errors in Spacy's sentence identification algorithm). 
\textit{TextTiling}, which decides the task structure through lexical overlap, performs poorly and does not appear to be a good option for this task. This is because recipe steps are not structured based on overlap, but rather in a sequence of sub-actions, which \textit{TextTiling} overlooks.

% supervised baselines
The \textit{CrossSeg} achieved a $P_k$ of $19.5$ which is already a significant improvement over the best unsupervised baseline which achieved $23.3$. This translates into an F1 score improvement from $73.8\%$ to $76.5\%$.
Regarding the \textit{DTS} models, the most solid fact that emerges from Table~\ref{tab_results_summary_2} is that, regardless of the backbone, our \textit{DTS} approach consistently outperforms all the baselines. 

In general, the results of the baselines illustrate the difficulty of the problem we are trying to solve. Moreover, there is a clear divide in terms of $P_k$ between previous baselines and the proposed \textit{DTS} framework, which is consistently below 20 $P_k$, highlighting the importance of capturing the relations between the task and the conversational steps.

\paragraph{Encoder vs Encoder-Decoder Backbones.}
% encoder vs encoder-decoder
Comparing the encoder-only model BERT~\cite{bert_original} with T5~\cite{t5_model} (Enc-only) or the full encoder-decoder (Enc-Dec) model, in situations with a comparable number of parameters, we see that \textit{T5} outperforms \textit{BERT}. This might be explained by the different pre-training approaches used in T5~\cite{t5_model}, which are better suited for our task.
% E-D models: encoder only vs encoder-decoder
Comparing the encoder-only (Enc-only) with the encoder-decoder (Enc-Dec) in the same models, we see an improvement in \textit{T5-Large}, but a decrease in performance in \textit{T5-Base}. This result implies that the use of the decoder part of \textit{T5} might not be necessarily needed for this particular task.

\paragraph{DTS Model Size Influence.}
% model size influence
Having established the performance range of \textit{DTS}, we examined the relationship between model size and performance. Results show that increasing the model size can bring improvements, in particular, from \textit{BERT-Base} to \textit{BERT-Large}, however, in the case of \textit{T5-Base} for \textit{T5-Large}, we notice an improvement in the Enc-Dec model and a decrease in performance in the encoder-only (Enc-only) model. Nonetheless, the best results by a significant margin in $\bm{P}_k$ and F1 are obtained with the largest model \textit{T5-3B} (Enc-only), showing that the use of larger models can bring an improvement as evidenced in ~\cite{t5_model, palm}.

\vspace{-5pt}
\paragraph{Main Takeaways.}
% finishing remarks
Results indicate that the proposed models are capable of capturing the intrinsic relations of the steps and extract them correctly when trained on high-quality conversationally structured task instructions. We also observe that our \textit{DTS}-Transformer approach gives the best results in this setting. A fundamental difference between \textit{DTS} and other supervised approaches is that it tackles the conversational recipe structuring task at a token-level granularity. As a consequence, it abstracts less information than previous approaches~\cite{text_segmentation_bert,transformer_squared}, such that at each Transformer layer, intermediate token embeddings are contextualized on the full-task sequence. Despite working at a finer granularity (token-level), \textit{DTS} is both faster to train and perform inferences. This makes it highly suited to be applied in a real setting, to structure tasks into conversational steps.

\subsubsection{Conversational Tasks Statistics}
The recipe task structuring results led us to further examine the resulting conversational steps statistics. These are shown in Table~\ref{tab_break_stats_2}, where we contrast the \# Steps and \# Tokens statistics with the human-annotated set. 
Specifically, we observe that Exact Match segmentation is higher (17.0\%) in the \textit{DTS T5-3B (Enc-only)} model due to its greater ability to capture the segmentation patterns. 
It is also interesting to note that all methods have a tendency to overestimate the number of steps.
Finally, for $\Delta$Steps$\leq1$ -- the percentage of examples where the model predicts less than one step of difference with the test set --  we see an equivalent performance within the supervised baselines.  

Overall, by examining these task structuring statistics, we observe that although the average number of steps (\# Steps column) is acceptable for most methods, when we look at the finer-grain statistics, we see that there is a non-trivial balance between step length, number of steps, and content of each step. Hence, it is not a sufficient condition to optimize a single statistic but rather a combination of these.

\begin{table*}[t]
\centering
\resizebox{\linewidth}{!}{%
\begin{tabular}{p{0.5cm}lccccccc}
%\hline
\toprule
& \textbf{} & \textbf{\# Steps} & \textbf{\# Tokens} & \textbf{Exact Match} & \textbf{= \# Steps} & \textbf{$+$ \# Steps} & \textbf{$-$ \# Steps} & \textbf{$\Delta$Steps$\leq 1$} \\ \midrule
\multicolumn{2}{l}{\multirow{1}{*}{\ \ \ \ \ \ \ \ \ \ Human Annotation}}           & \multirow{1}{*}{5.86}               & \multirow{1}{*}{19.21}              & \multirow{1}{*}{-}                   & \multirow{1}{*}{-}                   & \multirow{1}{*}{-}                     & \multirow{1}{*}{-}                    & \multirow{1}{*}{-}                    
\\\midrule
\multirow{4}{*}{\spheading[5em]{ \textbf{Method}} }
& Every$_1$          & 9.29              & 12.11              & 5.00\%                    & 5.33\%                & 94.67\%                 & 0.00\%                     & 24.00\%                   \\
& Text Tiling        & 6.32              & 17.80              & 7.00\%                    & 24.00\%                  & 49.33\%                 & 26.67\%                 & 58.67\%                \\ \cmidrule{2-9}
& CrossSeg      & 6.08              & 18.53              & 13.33\%                & 30.67\%               & 36.22\%                 & 33.11\%                 & 68.11\%                \\
& DTS T5-3B (Enc-only)          & 6.48              & 17.37              & 17.00\%                & 27.56\%               & 46.44\%                 & 26.00\%                 & 68.44\%                \\ \bottomrule
\end{tabular}%
}
\caption{Detailed conversational task structuring statistics for the ConvRecipes test set (human annotated). Exact Matches is the percentage of predictions exactly matching the ground-truth. (=, + and -) \# Steps represent the percentage of predictions that have equal, more, or less steps than the ground-truth. $\Delta$Steps$\leq 1$ indicates the percentage of times the difference between the \# Steps predicted and the ground-truth is $\leq 1$ step.}
\label{tab_break_stats_2}
\end{table*}

\begin{table}[htbp]
\centering
\resizebox{1.0\linewidth}{!}{%
\begin{tabular}{@{}lcc@{}}
\toprule
\textbf{}        & \textbf{Web-based} & \textbf{T5-3B (E-only)} \\ \midrule
Rating 1             & 18.0\%            & 3.3\%             \\
Rating 2             & 36.0\%            & 12.7\%            \\
Rating 3             & 18.7\%            & 20.7\%            \\
Rating 4             & 20.0\%            & 35.3\%            \\
Rating 5             & 7.3\%             & 28.0\%            \\ 
\midrule
Best                   & 14.0\%             & 86.0\%             \\
\midrule
Conv. Suitability & 2.63               & 3.72               \\ 
\bottomrule
\end{tabular}%
}
\caption{User study results comparing the original web-based segmentations with T5-3B (Enc-only) model predictions. (Conversation Suitability is given on a 1 to 5 scale.}
\label{tab_original_other_distribution_vs_model}
\end{table}

\subsubsection{User Evaluation}
To compare the model's performance to the original web-based instructions, we asked 6 annotators from the same pool of Section~\ref{sec_dataset} to annotate which segmentation was the best considering a conversational setting.

In total, 50 recipes were randomly selected from the test set in their original web-based format (i.e., without human annotations). 
These recipes were then compared to the predictions of the best model \textit{DTS T5-3B} (Enc-only).
Examples can be seen in Appendix~\ref{app_segmentation examples}.
For each recipe, we collected 3 annotations, resulting in an inter-rater agreement of 73\% w.r.t. binary preference. Additionally, the annotators were also asked to grade each segmentation (web and model) on a 1 to 5 Likert scale according to the suitability for a conversational agent.

Table~\ref{tab_original_other_distribution_vs_model} shows the results of the user evaluation. 
We observe a preference for the model's segmentation (86\%) since it was trained on a conversational-based data distribution which more accurately reflects the user's preference in this setting.
We also analyzed that the annotators had a preference for recipes with more segments 88\% of the time. Notwithstanding, it is important to note that breaking too often may result in a sub-optimal experience and in incomplete steps, as shown by the \textit{Every$_1$} baseline of Table~\ref{tab_results_summary_2}.

Considering the 1 to 5 rating of suitability for a conversational agent, the model's prediction scores were much higher (3.72) than the original recipes (2.63). These ratings further reinforce our hypothesis that the original recipes are not dialogue suited, and that the model is able to greatly increase the suitability of a recipe to a conversational-friendly format.

To conclude, these results indicate that the model is able to capture segmentation patterns, showing an ability to improve the suitability of a recipe for a conversational assistant.
This, in turn, brings advantages to the user experience by providing a grounded conversationally-suited segmentation.

\section{Conclusions}
\label{sec:conclusions}
In this paper, we proposed a methodology to tackle the problem of converting web/reading structured instructions into conversationally structured ones, using a task-grounded segmentation by considering the original task's steps.
In summary, the key contributions are as follows:

\paragraph{ConvRecipes Corpus.}
This corpus enables a better understanding of the problem. Its analysis showed that instructional text as it is presented online is not optimal for a conversational setting. 
    
\paragraph{Dialogue-Task Structurer (DTS).}
We proposed several methods that can effectively  
capture segmentation linguistic patterns.  
The best-performing method was a T5-3B (Enc-only) model, a token-level Transformer. 

\paragraph{Real-World Improvement.}
The user evaluation showcased the model's ability to improve over the original segmentation (86\%), which brings advantages in user experience in a conversational-assistant scenario.

\vspace{3mm}
For future work, we intend to assess how segmentation influences the user's perception of a recipe's quality and generalize our experiments to different domains such as DIY tasks and tutorials.

\section*{Acknowledgments} 
This work has been partially funded by the FCT project NOVA LINCS Ref. UIDP/04516/2020, by the Amazon Science - TaskBot Prize Challenge and the {CMU|Portugal}{} projects iFetch CMUP LISBOA-01-0247-FEDER-045920), and by the FCT Ph.D. scholarship grant UI/BD/151261/2021.
Any opinions, findings, and conclusions in this paper are the authors' and do not necessarily reflect those of the sponsors.

\bibliography{anthology,custom}
\bibliographystyle{acl_natbib}

\clearpage

\appendix

% \subsection{Dataset Metrics}
% \label{app_dataset_metrics}
% The main metric of the dataset is $P_k$~\cite{pk_metric}, which compares the predicted segmentation with the ground-truth labels, using the following formula:
% \begin{equation}
%     P_k = \sum_{1 \leq s \leq t \leq T} \mathbb{1}(\delta_{true}(s,t) \neq \delta_{hyp}(s,t)),
% \end{equation}
% where $T$ is a document with various sentences ($s$), and the function $\delta$ outputs 1 when sentences $s$ and $t$ belong to the same segment in the ground-truth and 0 otherwise. The function $\mathbb{1}(a \neq b)$ is equal to 1 when $a$ is different from $b$ and 0 otherwise.
% As in previous works~\cite{pk_metric,segbot,text_segmentation_bert}, we set the sliding window size $k$ to half of the document length divided by the total number of ground-truth segments. 
% In this setting a lower $P_k$ represents a better model. 
% To implement $P_k$, we used the reference implementation from NLTK~\cite{nltk}.

% Following~~\cite{segbot} we also consider as additional metrics Precision, Recall, and F-score defined as:
% \begin{equation}
%     Prec.=\frac{c}{h}, \ \ Rec.=\frac{c}{g}, \ \ \operatorname{\textit{F-score}} = \frac{2c}{g+h},
% \end{equation}
% where $g$ is the number of segments in the ground-truth, $h$ is the number of segments in the model's output, and $c$ is the number of correct segments in the model's output.

\section{Task Grounded Dialogue Segmentation Example in Conversation}
\label{example_dialog_table}
In Table~\ref{tab_recipe_in_context}, we show an example dialog between a user and a conversational assistant.
It is clear that segmenting a task is critical to managing the user's cognitive load, and allows for a better understanding of the recipe's multiple steps.

\begin{table}[ht]
\footnotesize
\centering
\caption{Example of a dialogue-suited segmented recipe in the context of a conversational task assistant.}
\label{tab_recipe_in_context}
\resizebox{\linewidth}{!}{%
\begin{tabular}{@{}p{0.7\linewidth}p{0.3\linewidth}@{}}
\toprule
\textbf{System}                                                                                                          & \textbf{User}               \\ \midrule
Hello! I am your cooking assistant! Try asking me things like: How to make lasagna? or How to bake a cake?                & I want to make a Tiramisu.   \\ \midrule
Sure, I found this very tasty: "Classic Tiramisu". Do you want to try it?                                                & Sure.                        \\ \midrule
Step 1. Separate eggs. Whisk egg yolks with confectioner’s sugar until pale and fluffy.                                  & How many eggs should I use? \\ \midrule
This recipe uses 2 eggs.                                                                                                 & Next.                        \\ \midrule
Step 2. Sure, let's keep going. Add Amaretto and mascarpone. Whisk until smooth.                                         & Next Step.                   \\ \midrule
Step 3. Beat egg whites with a pinch of salt until stiff peaks form. Slowly whisk in the remaining confectioner’s sugar. & Go ahead.                    \\ \midrule
Step 4. We are halfway through the task. Now, carefully fold beaten egg whites into mascarpone cream.                    & Next Step.                   \\ \midrule
Step 5. Combine espresso and remaining Amaretto in a shallow dish. Dip ladyfingers in espresso mixture and then layer soaked ladyfingers in bottom of the serving dish. &
  What are ladyfingers? \\ \midrule
Glad you ask! Ladyfingers are low-density, dry, egg-based, sweet sponge cake biscuits roughly shaped like large fingers. & Next Step.                   \\ \midrule
Step 6. Cover with a layer of mascarpone cream and top with a fine layer of grated chocolate.                            & I'm done.                    \\ \midrule
Step 7. Repeat layering process until all ingredients are used up. Finish up with a layer of mascarpone cream and dust with unsweetened cocoa powder. &
  Next. \\ \midrule
Step 8. This is the last step of this recipe. Refrigerate for at least 3 hours before serving.                            & Next.                        \\ \midrule
Excellent work, we just finished Classic tiramisu together! Hope you had fun with me today!                              & Goodbye!                           \\ \bottomrule
\end{tabular}%
}
\end{table}

\section{Conversational Steps Traits Description}
\label{app_traits_description}
The following traits were classified in a 1 to 5 Likert scale considering their importance in a task-guiding setting:
\begin{itemize}
    \item \textbf{Step Complexity} - Was the complexity of the step an important factor to segment?
    \item \textbf{Step Clarity} - Was the clarity of a step important, i.e. information was clear, organized and well delimited in each step?
    \item \textbf{Step Length and Number of Steps} - Did the length of the steps and the total number of steps influence your decision? 
    \item \textbf{Ability to Parallelize Tasks} - Were you concerned that the segmentation should allow the user to parallelize multiple tasks/steps?
    \item \textbf{Naturalness} - Was the naturalness of each step important for your decision? i.e., sounded close to a person explaining the task?
\end{itemize}

\section{Model Training and Hyperparameters}
\label{app_implementation_details}
We trained every model for 20 epochs and evaluated in the test set the model with the best performance in the validation set in terms of F-Score.
We used a batch size of 16, a learning rate of  $5^{-5}$, and the Adam optimizer~\cite{adam_optimizer}. All models were trained on a single NVIDIA A100 GPU, except for T5-3B (Enc-only), which used 4 NVIDIA A100 GPUs. 
%The average training time for the models was: 10, 20, 35 minutes for DTS-Transformer models with 110, 220 and 340 million parameters, respectively. For DTS models with 770 million (T5-Large (Enc-Dec)) and 1.5 million (T5-3B (Enc-only)) the training time was 1 hour.
We also highlight that the \textit{DTS} models are faster to train compared to the cross-segment model~\cite{text_segmentation_bert}%, which takes about 1 hour with 110 M parameters
, due to each training sample only predicting one segmentation, instead of all segments as in DTS.

\subsection{Attention Visualization}
\label{app_attention_visualization}
\begin{figure*}[ht!]%
    \centering
    {{\includegraphics[width=0.30\linewidth]{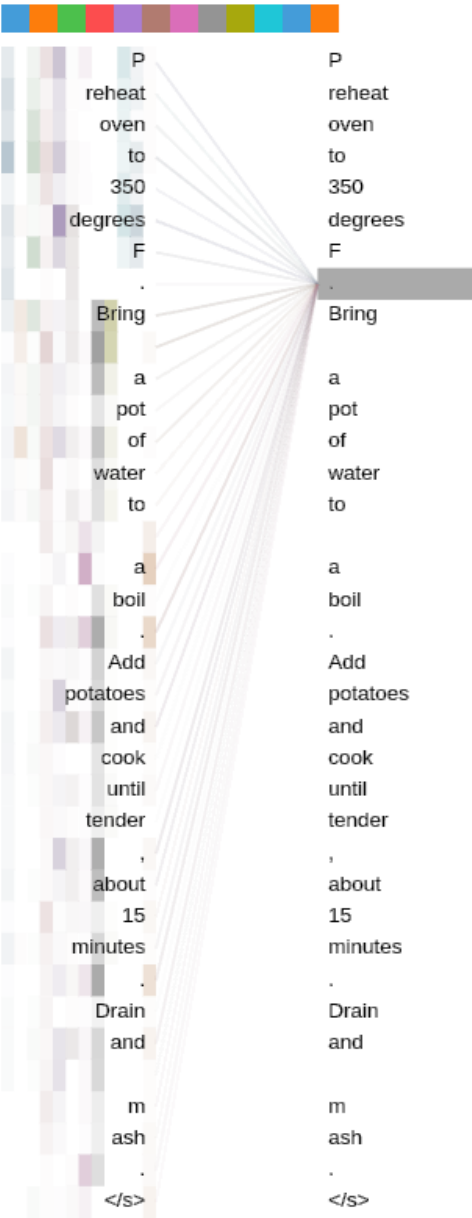} }}%
    \hspace{6mm}
    {{\includegraphics[width=0.30\linewidth]{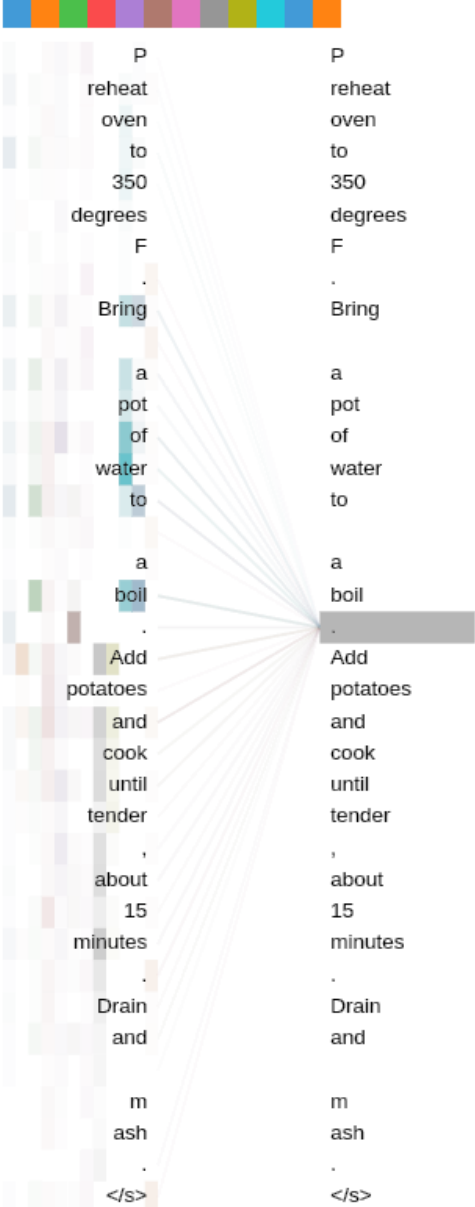} }}%
    \hspace{6mm}
    {{\includegraphics[width=0.30\linewidth]{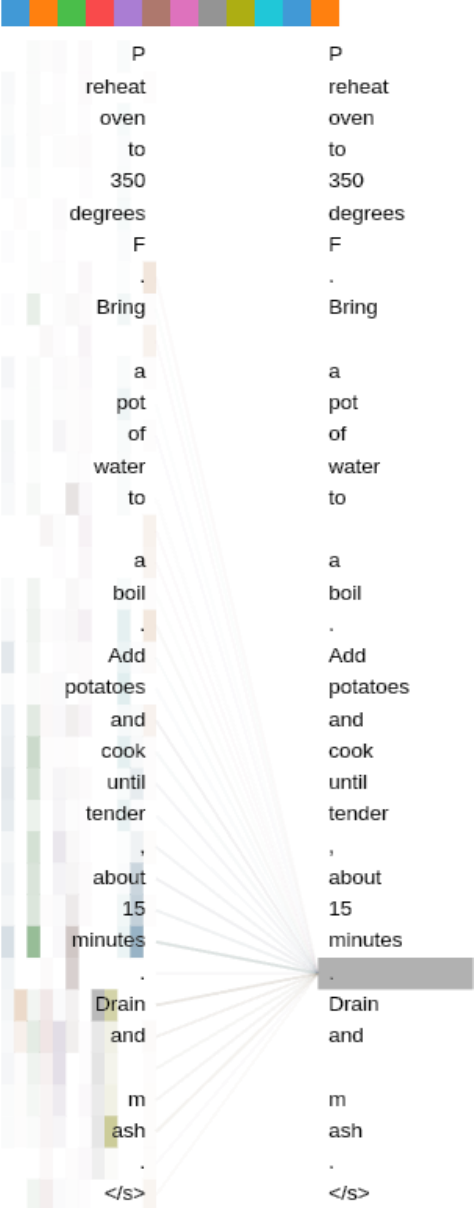} }}%
    % {{\includegraphics[width=0.23\linewidth]{images/attention_example_4.pdf} }}%
    \caption{Attention values for possible segmentation points on the recipe "Mashed Potatoes" in the last layer of T5-Base (Enc-only), created using~\cite{bertviz}.}%
    \label{fig_attention_visualization}%
\end{figure*}

Given the self-attention mechanism of Transformer architectures~\cite{vaswani_attention}, in Figure~\ref{fig_attention_visualization}, we show the attention given to each token on a possible segmentation point using the DTS T5-Base (Enc-only) model.
It is interesting to note that each possible segment attends to its neighboring sentences and particularly focuses on actionable words such as verbs (e.g., "boil"), and common segment terminating tokens, such as mentions to temperatures (e.g. "305 degrees") and time (e.g. "15 minutes").
This shows that the model is attending to the most relevant tokens of the recipe, which in turn, leads to a more insightful segmentation based on the conversational aspects learned during fine-tuning.

\section{Model Output Examples}   % case studies
\label{app_segmentation examples}
Table~\ref{tab_example_outputs} shows examples comparing the original web version of a recipe with the output of the T5-3B (Enc-only) DTS model.
As we can observe, the model outputs a larger number of breaks complementing our findings in this conversational scenario. 
The model is also able to keep a notion of the sub-task being performed, for instance in Example-3 Step-4, the model does not segment the sentences into multiple steps.

% \usepackage{graphicx}
\begin{table*}[htbp]
\renewcommand{\arraystretch}{1.1}
\small
\centering
\resizebox{\textwidth}{!}{%
\begin{tabular}{p{0.50\textwidth}p{0.50\textwidth}}
\toprule
\textbf{Example 1 - \textit{Soy Garlic Steak}} (Web) & \textbf{Model Output} \\ \midrule
\begin{tabular}[t]{@{}p{0.5\textwidth}@{}}\textbf{1.} In a small bowl, mix vegetable oil, soy sauce, vinegar, ketchup, and crushed garlic. Place flank steak in a large resealable plastic bag. Pour the marinade over steak. Seal, and marinate in the refrigerator at least 3 hours.\\ \\ \textbf{2.} Preheat grill for high heat.\\ \\ \textbf{3.} Oil the grill grate. Place steaks on the grill, and discard marinade. Cook for 5 minutes per side, or to desired doneness.\end{tabular} & \begin{tabular}[t]{@{}p{0.5\textwidth}@{}}\textbf{1.} In a small bowl, mix vegetable oil, soy sauce, vinegar, ketchup, and crushed garlic.\\ \\ \textbf{2.} Place flank steak in a large resealable plastic bag. Pour the marinade over steak. Seal, and marinate in the refrigerator at least 3 hours.\\ \\ \textbf{3.} Preheat grill for high heat.\\ \\ \textbf{4.} Oil the grill grate. Place steaks on the grill, and discard marinade. Cook for 5 minutes per side, or to desired doneness.\end{tabular} \\ \midrule
\textbf{Example 2 - \textit{Blueberry Yogurt Pops}} (Web) & \textbf{Model Output} \\ \midrule
\begin{tabular}[t]{@{}p{0.5\textwidth}@{}}\textbf{1.} Combine all ingredients in blender. Cover; blend on high speed 15 to 20 seconds or until smooth.\\ \\ \textbf{2.} Pour into 8 frozen pop molds (2.5 ounces to 3 ounces each). Cover, insert craft sticks, and freeze for 2 hours or until completely firm.\\ \\ \textbf{3.} To serve, dip outsides of molds into warm water to loosen.\end{tabular} & \begin{tabular}[t]{@{}p{0.5\textwidth}@{}}\textbf{1.} Combine all ingredients in blender. Cover; blend on high speed 15 to 20 seconds or until smooth.\\ \\ \textbf{2.} Pour into 8 frozen pop molds (2.5 ounces to 3 ounces each).\\ \\ \textbf{3.} Cover, insert craft sticks, and freeze for 2 hours or until completely firm.\\ \\ \textbf{4.} To serve, dip outsides of molds into warm water to loosen.\end{tabular} \\ \midrule
\textbf{Example 3 - \textit{Quinoa Salad with Roasted Yams}} (Web) & \textbf{Model Output} \\ \midrule
\begin{tabular}[t]{@{}p{0.5\textwidth}@{}}\textbf{1.} Preheat oven to 350 degrees F (175 degrees C). Line a baking sheet with aluminum foil; add yams.\\ \\ \textbf{2.} Bake in the preheated oven until yams are tender and wrinkled at the edges, about 20 minutes. Cool to room temperature, about 15 minutes\\ \\ \textbf{3.} Bring water to a boil in a large saucepan. Add quinoa, stirring once; return to boil. Cook uncovered until water is absorbed, 10 to 12 minutes. Strain, shaking the sieve well to remove all moisture. Transfer to a mixing bowl.\\ \\ \textbf{4.} Stir cucumbers, yams, parsley, olive oil, onion, lemon juice, red wine vinegar, salt, and pepper into the quinoa. Garnish with endive spears.\end{tabular} & \begin{tabular}[t]{@{}p{0.5\textwidth}@{}}\textbf{1.} Preheat oven to 350 degrees F (175 degrees C). \\ \\ \textbf{2.} Line a baking sheet with aluminum foil; add yams.\\ \\ \textbf{3.} Bake in the preheated oven until yams are tender and wrinkled at the edges, about 20 minutes. Cool to room temperature, about 15 minutes\\ \\ \textbf{4.} Bring water to a boil in a large saucepan. Add quinoa, stirring once; return to boil. Cook uncovered until water is absorbed, 10 to 12 minutes. \\ \\ \textbf{5.} Strain, shaking the sieve well to remove all moisture. Transfer to a mixing bowl.\\ \\ \textbf{6.} Stir cucumbers, yams, parsley, olive oil, onion, lemon juice, red wine vinegar, salt, and pepper into the quinoa. Garnish with endive spears.\end{tabular} \\ \bottomrule
\end{tabular}%
}
\caption{Examples comparing original web recipe and the T5-3B (Enc-only) DTS model outputs.}
\label{tab_example_outputs}
\end{table*}

\end{document}